\documentclass[usletter,conference]{IEEEtran}
\ifCLASSINFOpdf
  \usepackage[pdftex]{graphicx}
\else
  \usepackage[dvips]{graphicx}
\fi

\usepackage{amssymb}
\usepackage{amsmath}
\usepackage{wasysym}

\usepackage{hyperref}

\begin{document}
\title{FreSaDa: A French Satire Data Set for Cross-Domain Satire Detection}

\author{\IEEEauthorblockN{Radu Tudor Ionescu\IEEEauthorrefmark{1},
Adrian Gabriel Chifu\IEEEauthorrefmark{2}}
\IEEEauthorblockA{
\IEEEauthorrefmark{1}Deparment of Computer Science and Romanian Young Academy,\\
University of Bucharest, Romania\\
Email: raducu.ionescu@gmail.com
}
\IEEEauthorblockA{
\IEEEauthorrefmark{2}
LIS UMR CNRS 7020,\\
Aix-Marseille Université/Université de Toulon, France\\
Email: adrian.chifu@univ-amu.fr}}


\maketitle

\begin{abstract}
In this paper, we introduce FreSaDa, a \textbf{Fre}nch \textbf{Sa}tire \textbf{Da}ta Set\footnote{\url{https://github.com/adrianchifu/FreSaDa}}, which is composed of 11,570 articles from the news domain. In order to avoid reporting unreasonably high accuracy rates due to the learning of characteristics specific to publication sources, we divided our samples into training, validation and test, such that the training publication sources are distinct from the validation and test publication sources. This gives rise to a cross-domain (cross-source) satire detection task. We employ two classification methods as baselines for our new data set, one based on low-level features (character n-grams) and one based on high-level features (average of CamemBERT word embeddings). As an additional contribution, we present an unsupervised domain adaptation method based on regarding the pairwise similarities (given by the dot product) between the training samples and the validation samples as features. By including these domain-specific features, we attain significant improvements for both character n-grams and CamemBERT embeddings.
\end{abstract}


%
\IEEEpeerreviewmaketitle

\section{Introduction}

Satirical (or ironical) text detection is a preliminary step towards building conversational systems and robots that are capable of understanding and producing satirical text, during their interaction with humans. Indeed, inaccurate detection of satire (or irony) might lead to wrong actions taken by such conversational systems or robots when these AI-based systems literally interpret satirical comments of humans as commands. Therefore, satire detection is an important topic in computational linguistics. Up until now, satire detection has been studied in various languages such as English \cite{Reganti-ICDMW-2016}, German \cite{McHardy-NAACL-2019} and Spanish \cite{Barbieri-PLN-2015}. 
Even more languages were subject to irony or sarcasm detection research, including Arabic \cite{Karoui2017}, Dutch \cite{Liebrecht2013}, Chinese \cite{Jia2019}, Italian \cite{Giudice-EVALITA-2018}, to name only a few.

In this paper, we focus on satire detection in French, proposing FreSaDa, a \textbf{Fre}nch \textbf{Sa}tire \textbf{Da}ta Set of news articles collected from regular and satirical publication sources. An important aspect of our study is to underline that there is a clear difference between satire and fake news detection \cite{Meel-ESA-2019,Perez-COLING-2018,Sharma-TIST-2019}. One difference is that fake news is not necessarily written in a satirical style. The most dangerous fake news articles are those written to seem as realistic as possible, their main purpose being that of misinforming people. Satirical news can be easily identified by humans as fake, their purpose being to cause laughter among readers. Hence, fake news detection and satirical news detection have completely different goals. We therefore consider studies on fake news detection as being outside the scope of this work.

Along with our new data set, FreSaDa, we propose two baseline methods to be used as reference in future work. The first baseline is a shallow approach based on low-level features, namely character n-grams. The second baseline is a deep method based on CamemBERT word embeddings \cite{Martin-Arxiv-2019}. We compare the two methods in two binary classification settings: full news article classification into \emph{regular} versus \emph{satire}, and headline (title) classification into \emph{regular} versus \emph{satire}. The model based on CamemBERT embeddings attains better results on full news articles, while the model based on character n-grams attains superior performance on the more challenging headline classification task.

In summary, our contribution is twofold:
\begin{itemize}
\item We introduce a novel data set of French news articles collected from regular and satirical publication sources, which allows us to perform cross-source satire detection in various settings.
\item We propose a novel, effective and generic unsupervised domain adaptation method, which brings significant improvements for both character n-grams and CamemBERT embeddings.
\end{itemize}

The rest of this paper is organized as follows. The related work is presented in Section~\ref{sec_RW}. Our data set is described in Section~\ref{sec_Dataset}. The baselines as well as our unsupervised domain adaptation method are presented in Section~\ref{sec_Method}. We present experiments and results on full news articles and headlines in Sections~\ref{sec_Exp_full} and~\ref{sec_Exp_titles}, respectively. We also discuss the most discriminative features of the two baselines in Section~\ref{sec_Analysis}. Finally, we draw our conclusions in Section~\ref{sec_Conclusion}.

\section{Related Work}
\label{sec_RW}

\subsection{Satire Detection in English}

In \cite{Burfoot-ACL-2009}, the authors tackle the task of discriminating between satirical and regular news, proposing the first English corpus for satire detection with 4,000 regular news articles and 233 satirical news articles. 

Frain and Wubben \cite{Frain-LREC-2016} have created a balanced multi-domain (politics, entertainment and technology) English data set for the same task (1,706 satirical articles and 1,705 regular articles). The regular news articles are collected from websites such as Reuters, CNET, CBS News, while the satirical articles come from  websites such as Daily Currant, DandyGoat, EmpireNews, NewsBiscuit, NewsThump, SatireWire and The Spoof. The authors have designed three experiments, based on the type of the employed features: (1) unigram and bigram BOW-based features, (2) manually crafted features such as profanity, punctuation, sentiment polarity and human-centeredness and (3) the combination of the two aforementioned features. 

In \cite{Goldwasser-TACL-2016}, the authors introduce \textit{ComSense}, a latent variable model for satire detection. They argue that this task is inherently a common-sense reasoning task, rather than a traditional text classification task. The employed data set is the one proposed in \cite{Burfoot-ACL-2009}.

Li et al.~\cite{Li-NLP4IF-2020} have proposed a multi-modal (text and image) approach based on the ViLBERT model \cite{Lu2019}, for the task of satire detection. They also propose a data set of thumbnail images and headlines of regular (6,000 samples) and satirical (4,000 samples) news articles, in English. They fine-tune ViLBERT on the data set and train a CNN that uses an image forensics technique. 

Ravi and Ravi \cite{Ravi-KBS-2017} have proposed an ensemble feature selection method followed by a framework with several classifiers to automatically detect satire, sarcasm and irony found in news and customer reviews. They have used three data sets, two for satire and one for irony. The first satiric data set is the one from \cite{Burfoot-ACL-2009}, while the second has been built by the authors themselves (1,272 regular and 393 satiric news articles). 

Yang et al.~\cite{Yang-EMNLP-2017} have considered paragraph-level linguistic features to unveil the satire through a neural network classifier with attention. They have investigated the difference between paragraph-level features and document-level features, and have analyzed them on a large satirical news data set, in English. The 16,249 satirical news articles have been collected from 14 websites and the 168,780 regular news articles have been collected from several major news outlets, such as CNN, DailyMail, The Guardian, among others. 

\subsection{Satire Detection in Other Languages}

The works discussed so far~\cite{Burfoot-ACL-2009,Frain-LREC-2016,Goldwasser-TACL-2016,Li-NLP4IF-2020,Ravi-KBS-2017,Yang-EMNLP-2017} study satire detection on English news. There are considerably less studies on satire in other languages. 

The authors of \cite{Saadany-RDSM-2020} have aimed to identify the linguistic properties of Arabic fake news with satirical content through a series of exploratory analyses. They have concluded that Arabic satirical news has distinguishing features on the lexico-grammatical level, with respect to the regular news. In order to conduct their study, the authors have built an Arabic data set containing 3,185 satirical articles from two websites and 3,710 regular articles from three websites. 

In \cite{Tocoglu-ID-2019}, the authors have presented an approach based on machine learning to detect satire in Turkish news articles. They have employed three kinds of features to model lexical information: unigrams, bigrams and trigrams. Term-frequency, term-presence and TF-IDF based schemes have also been considered. As classifiers, Naïve Bayes, SVM, logistic regression and C4.5 algorithms have been studied. The data set consists of 500 satirical and 500 regular Turkish news articles.

The authors of \cite{McHardy-NAACL-2019} have proposed an LSTM with attention model for satire detection, with an adversarial component to control for the confounding variable, namely the publication source. They have built a German data set containing 320,219 regular articles 9,643 satirical articles, collected from 11 websites. This is a highly unbalanced data set.


\subsection{Satire Detection in French}

To our knowledge, there are only a few studies on French satire detection \cite{Guibon-CICLing-2019,Liu-ICCCN-2019}. Guibon et al.~\cite{Guibon-CICLing-2019} compared several automatic approaches for fake news detection, based on statistical text analysis on the vaccination fake news data set provided by the Storyzy company. The data set was split into three parts: English, French and YouTube, with the French part being formed of 705 samples for training and 236 samples for testing.
Their CNN architecture worked better for the discrimination of broader classes (fake versus trusted), while the gradient boosting decision tree based on feature stacking obtained better results for satire detection. They showed that efficient satire detection can be achieved using merged embeddings and a specific model, at the cost of lower performance on broader classes. They also merged redundant information, in order to better distinguish satirical news from fake news and trusted news. With respect to the work of Guibon et al.~\cite{Guibon-CICLing-2019}, we mention that our data set is larger by an order of magnitude.

Liu et al.~\cite{Liu-ICCCN-2019} tackled the issue of automatically detecting satirical and false news in French. They provided a French satire data set containing 5,682 French-language articles from six different sources (four satirical and two regular). They have also introduced some baseline methods that discriminate between satirical and regular news, based on Logistic Regression, neural networks, Support Vector Machines, Random Forest and Na\"{i}ve Bayes. Their research is the closest to ours, but the data set we propose is larger and the baseline methods are more advanced.  

Unlike Guibon et al.~\cite{Guibon-CICLing-2019} and Liu et al.~\cite{Liu-ICCCN-2019}, in our approach, the examples in training and testing are issued from different publication sources. This enables us to perform a more realistic evaluation, preventing machine learning models from taking decisions based on features specific to the publication source, e.g.~the style of writers from certain publication sources. The task that we propose can be viewed as cross-source or cross-domain satire detection. 
We note that cross-source satire detection has also been studied in German \cite{McHardy-NAACL-2019}. In their study, McHardy et al.~\cite{McHardy-NAACL-2019} showed that the adversarial component can help the neural model in learning to pay attention to the linguistic properties of satire. Instead of adversarial training, we propose an unsupervised domain adaptation method based on regarding the pairwise similarities (given by the dot product) between the training samples and the validation samples as features. By including these domain-specific features, we attain significant improvements for the proposed baseline methods.

%

\section{Data Set}
\label{sec_Dataset}

\begin{table}[!t]
\caption{The number of regular and satirical news articles (\#samples) and the corresponding number of tokens (\#tokens) contained in the training, validation and test sets of FreSaDa.}
\begin{center}
\begin{tabular}{|l|r|r|r|r|}
\hline
Set 						& \multicolumn{2}{|c|}{Regular}    & \multicolumn{2}{|c|}{Satirical}\\
\cline{2-5}
     						& \#samples		&	\#tokens	    & \#samples		&	\#tokens\\

\hline
\hline
Training				    & 4,221 		& 2,247,102            & 4,495 		& 1,518,917\\
Validation					& 713			&
289,934 & 714		    & 353,813\\
Test					    & 714			& 283,768            & 713		    & 344,159\\
\hline
Total						& 5,648			& 2,820,804           & 5,922			& 2,216,889\\
\hline
\end{tabular}
\end{center}
\label{tab_stats}
\end{table}

To build the data set, we collected publicly available text samples from four French news websites, two of them being focused on satirical news. From the beginning, we decided to separate the publication sources between training, validation and test. Without being able to find three publication sources of French satire with enough data samples, we decided to use the same publication source for validation and testing. After we collected the data samples from the satire publication sources, we proceeded by collecting a matching number of regular news articles. The samples were collected from the same time period. We used two publication sources of regular news, in order to keep the separation of sources between training versus validation and test. In the end, we gathered 5,648 regular news samples and 5,922 satirical news samples. We used stratified sampling to divide the collected news articles into training, validation and test. In Table~\ref{tab_stats}, we present the number of satirical and regular news articles inside each subset (train, validation and test), along with the number of tokens. The training set contains 8,716 samples, while the validation and the test sets contain 1,427 samples each. With a total of 11,570 news articles and over five million tokens, FreSaDa is the largest corpus of its kind. We notice that our data set is well-balanced with respect to the sample distribution per news type (satirical versus regular). We also note that, in order to obtain the final text samples, we removed all HTML tags and kept only the article's title (headline) and textual content. We concatenated each title with the body of the corresponding article, leading to an average length of 435 tokens per text sample. We also create a more challenging benchmark based entirely on headlines. Hence, we consider two possible tasks on FreSaDa:
\begin{itemize}
    \item Cross-domain binary classification of full news articles into \emph{regular} versus \emph{satirical} examples.
    \item Cross-domain binary classification of headlines into \emph{regular} versus \emph{satirical} examples.
\end{itemize}

It is perhaps important to underline that our data set is intended for nonprofit educational purposes and not for commercial use. Since the collected news are in the public web domain, the noncommercial licensing follows the guidelines of the EU Copyright Directive 790/19\footnote{\url{https://eur-lex.europa.eu/eli/dir/2019/790/oj}}.


\section{Methods}
\label{sec_Method}

\subsection{String Kernels}

A simple language-independent and linguistic-theory-neutral approach consists in interpreting text samples as sequences of characters (strings) and using character n-grams as features. As the number of character n-grams is usually much higher than the number of samples, representing the text samples as feature vectors may require lots of space. String kernels \cite{Butnaru-VarDial-2018,Cozma-ACL-2018,Ionescu-VarDial-2017,Ionescu-EMNLP-2014,Gimenez-EACL-2017,Ionescu-COLI-2016} provide an efficient way to avoid storing and using the feature vectors (primal form), by representing the data though a kernel matrix (dual form). Each component $K_{ij}$ in a kernel matrix represents the similarity between data samples $x_i$ and $x_j$. As similarity (kernel) function, in our experiments, we consider either the presence bits string kernel (PBSK) \cite{Popescu-BEA8-2013} or the histogram intersection string kernel (HISK) \cite{Ionescu-EMNLP-2014}. For two strings $x_i$ and $x_j$ over a set of characters $S$, HISK is defined as follows:
\begin{equation}
k^{\cap}(x_i, x_j)=\sum\limits_{g \in S^n} \min \lbrace \mbox{\#}(x_i, g), \mbox{\#}(x_j, g) \rbrace ,
\end{equation}
where $\mbox{\#}(x,g)$ is a function that returns the number of occurrences of n-gram $g$ in $x$, and $n$ is the length of n-grams. PBSK is defined in a similar way, just by changing the function $\mbox{\#}(x,g)$ to return 1 whenever the number of occurrences of n-gram $g$ in $x$ is greater than 1.

\subsection{Average of CamemBERT Embeddings}

In order to build a stronger baseline based on high-level features, we consider CamemBERT \cite{Martin-Arxiv-2019}, a state-of-the-art language model for French. Following the success of BERT \cite{Devlin-NAACL-2019} and RoBERTa \cite{Liu-Arxiv-2019}, Martin et al.~\cite{Martin-Arxiv-2019} trained CamemBERT on the French version of the OSCAR corpus \cite{Suarez-CMLC-2019}, using the same neural architecture as RoBERTa, i.e.~CamemBERT is a multi-layer bidirectional transformer \cite{Vaswani-NIPS-2017}. CamemBERT produces 768-dimensional embeddings of words. We pass the news articles through CamemBERT, obtaining 768-dimensional vectors for every token. To obtain the final document-level representations, we simply average the word embeddings for each document in detriment of more complex frameworks \cite{Fu-ESA-2018,Ionescu-KES-2017,Ionescu-NAACL-2019}, as suggested by Shen et al.~\cite{Shen-ACL-2018}. 

It is important to note that we did not have the computational resources to fine-tune CamemBERT, which would likely produce better results. However, since we are presenting a new data set, our main goal is not to saturate FreSaDa by reporting outstanding performance levels, but only to provide some strong baselines for future comparison.

\subsection{Unsupervised Domain Adaptation}

We propose a novel, simple and generic domain adaptation method based on using unlabeled samples from the target domain, e.g.~samples taken from the validation set. Our unsupervised domain adaption method is based on two simple steps. Let $X=\{x_i \in \mathbb{R}^p\;|\;i = \overline{1,m} \}$ be a training set from the source domain and $Z=\{z_i \in \mathbb{R}^p\;|\;i = \overline{1,r} \}$ be an unlabeled validation set from the target domain. In the first step, we compute the dot product between each training feature vector $x_i$ and each validation feature vector $z_j$, as follows:
\begin{equation}
v_j = \langle x_i, z_j \rangle, \forall j = \overline{1,r}.
\end{equation}

In the second step, each training feature vector $x_i$ is concatenated with the vector $v_j$ of dot products between $x_i$ and every validation vector $z_j$. Naturally, the dimension of each vector of dot products is equal to the size of $Z$. With the notations defined above, the new training feature vectors have $p+r$ components. With these new features, a classifier can now assign weights that reflect how similar a training example is with the validation examples. In this way, the classifier is given the chance to rely on training samples that are more similar to samples from the target domain, thus adapting itself to the target domain.

\subsection{Learning Models}

In the learning stage, we employ a linear classifier that allows us $(i)$ to input either feature vectors or pre-computed kernel matrices and $(ii)$ to easily determine discriminative features in order to explain the predicted labels. Our first choice is to employ Support Vector Machines (SVM) \cite{Cortes-ML-1995}, a model that finds a hyperplane which separates the training samples by maximizing the margin. The margin is optimized with respect to the data points that are closest to the hyperplane, which are known as \emph{support vectors}. A similar approach that relies on all data points to find the separating hyperplane is Ridge Regression (RR) \cite{Hoerl-Technometrics-1970}, also known as linear regression with $l_2$ regularization. For PBSK and HISK, we employ the dual version, known as Kernel Ridge Regression (KRR) \cite{Saunders-ICML-1998}, which enables us to use pre-computed kernels. For CamemBERT embeddings, it is not necessary to use the kernel version of RR. Regardless of the data representation, in order to apply RR and KRR on our classification task, we need to apply the sign function on the predicted scores, transforming them into class labels from the set $\{-1, 1\}$.

\section{Experiments on Full News Articles}
\label{sec_Exp_full}

\begin{figure}[!t]
\begin{center}
\includegraphics[width=0.76\linewidth]{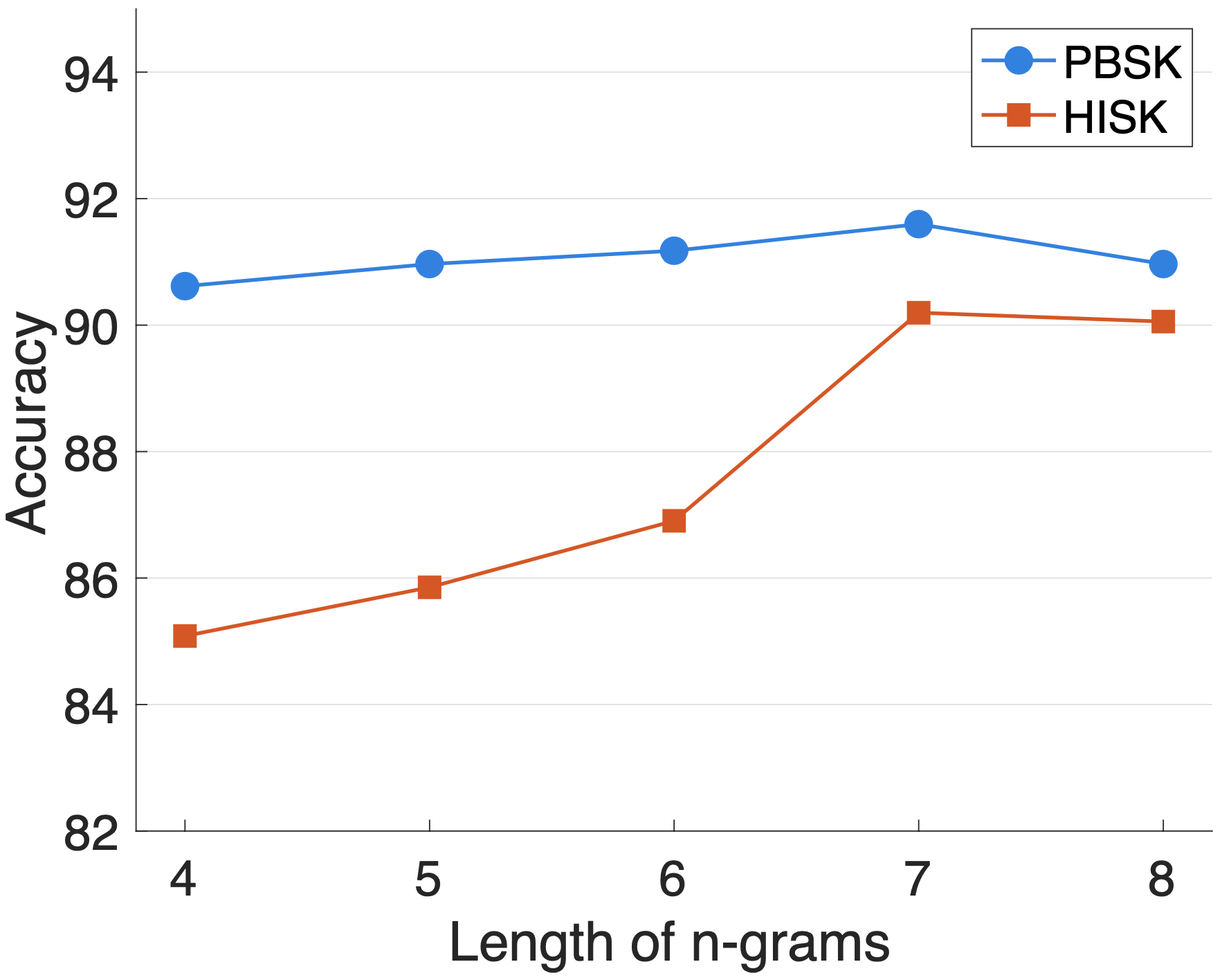}
\end{center}
\caption{Validation accuracy rates on full news articles for PBSK and HISK with character n-grams in the range $\{4,5,6,7,8\}$, respectively. Best viewed in color.}
\label{fig_ngrams_plot}
\end{figure}

\begin{figure}[!t]
\begin{center}
\includegraphics[width=0.78\linewidth]{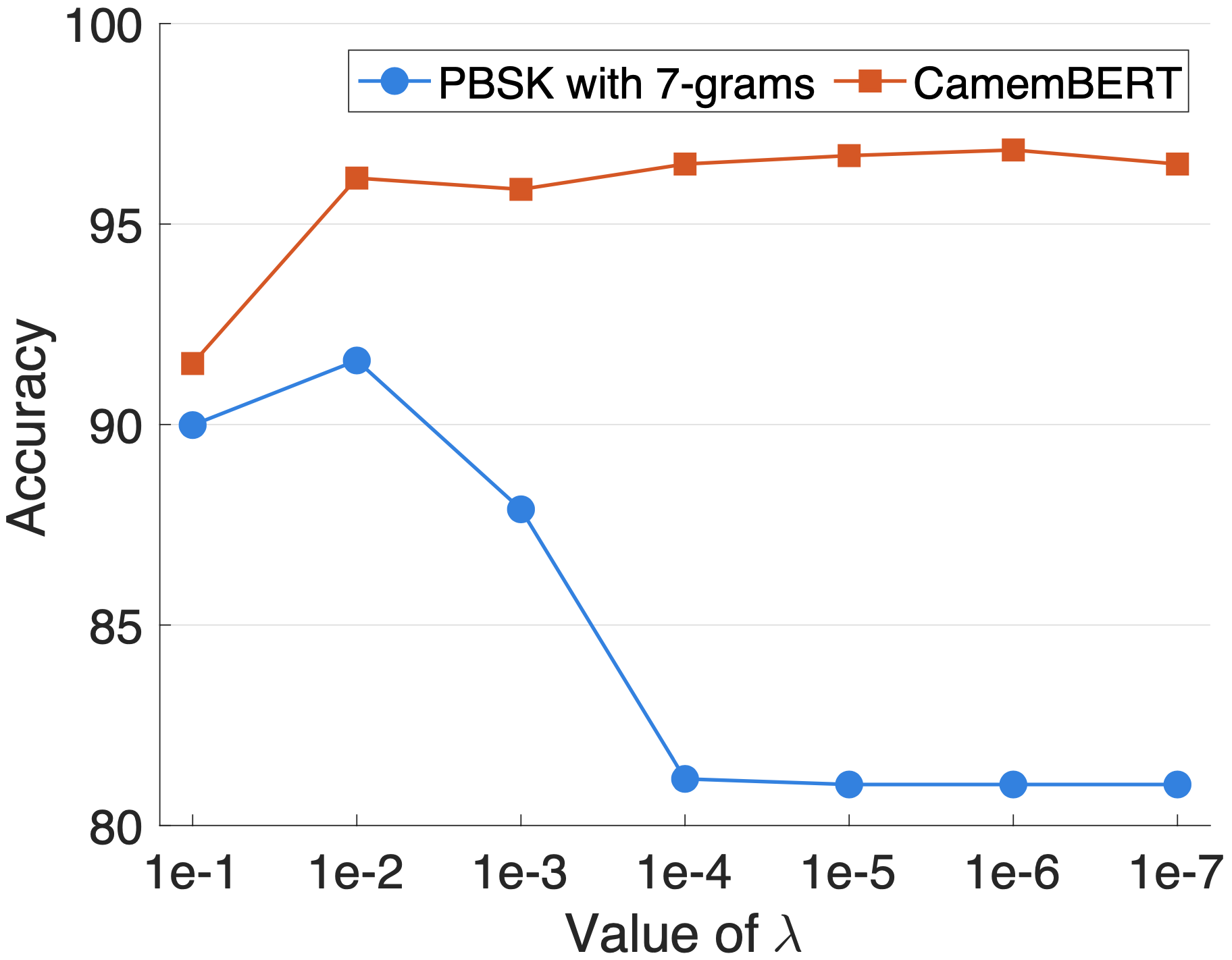}
\end{center}
\caption{Validation accuracy rates on full news articles for PBSK with character 7-grams and for CamemBERT embeddings, with values of $\lambda$ between $10^{-1}$ and $10^{-7}$, where $\lambda$ controls regularization in (Kernel) Ridge Regression. Best viewed in color.}
\label{fig_lambdas_plot}
\end{figure}

\subsection{Hyperparameter Tuning}

In a set of preliminary experiments with string kernels, we compared SVM and KRR, obtaining consistently better results with the latter model. The fact that KRR attains superior results to SVM has also been observed in a previous work \cite{Ionescu-COLI-2016} on text classification. Hence, we decided to employ KRR in the following experiments.

For PBSK and HISK, we considered character n-grams of length $n \in \{4,5,6,7,8\}$ as features. In Figure~\ref{fig_ngrams_plot}, we show the validation accuracy rates for each n-gram length, using both string kernel methods. We observe that the peak accuracy rates on full news articles are obtained when the n-gram length is $7$. It is important to note that we also tried PBSK based on 9-grams and 10-grams, confirming that the performance continues to drop. Based on the reported results, we opted for PBSK based on 7-grams in the subsequent experiments. For CamemBERT, we did not have to tune any parameters. In the learning stage, we employed the Ridge Regression classifier, irrespective of the data representation (PBSK, HISK or CamemBERT). 
However, the regularization parameter $\lambda$ of (Kernel) Ridge Regression was tuned on the validation set, using values in the set $\{10^{-1}, 10^{-2}, 10^{-3}, 10^{-4}, 10^{-5}, 10^{-6}, 10^{-7} \}$, for each data representation (PBSK and CamemBERT). In Figure~\ref{fig_lambdas_plot}, we show the validation accuracy rates for each value of $\lambda$ using either PBSK with 7-grams or the average of CamemBERT word embeddings. Considering the models with optimal regularization, for KRR based on PBSK, we opted for $\lambda=10^{-2}$, while for RR based on CamemBERT, we opted for $\lambda=10^{-6}$. Each time we integrated our unsupervised domain adaptation method into a classification model, we repeated the tuning of $\lambda$ for optimal results.

\begin{table}[!t]
\caption{Validation and test accuracy rates of (Kernel) Ridge Regression using PBSK or CamemBERT representations on full news articles. Results are reported with and without unsupervised domain adaptation (DA). Results of domain adapted methods marked with $\dagger$ are significantly better than the corresponding baseline, according to a paired McNemar's test \cite{Dietterich-NC-1998} performed at the significance level $0.05$.}
\begin{center}
\begin{tabular}{|l|c|c|}
\hline
Representation 				& Validation            & Test\\
\hline
\hline
PBSK				        & 91.60\%               & 91.17\% \\
CamemBERT			        & 96.85\%		        & 96.50\% \\
\hline
PBSK + DA                   & 92.86\%$^\dagger$	    & 93.34\%$^\dagger$ \\
CamemBERT + DA	            & 97.06\%		        & 97.48\%$^\dagger$ \\           
\hline
\end{tabular}
\end{center}
\label{tab_results_FreSaDa}
\end{table}

\subsection{Results}

In Table \ref{tab_results_FreSaDa}, we present the validation and the test results on full news articles from FreSaDa, using the proposed baselines. Considering the results without domain adaptation, we notice that the average of CamemBERT word embeddings attains much better results than the representation given by character n-grams (PBSK). This probably indicates that accurate satire detection requires high-level semantic features. Nonetheless, given that both methods yield accuracy rates above $90\%$, we conclude that satire detection in full news articles is not a very difficult task. Indeed, it seems that both models find enough discriminative clues in the full-length news articles.

Considering the results with domain adaptation, we notice significant performance gains on the test set for both PBSK and CamemBERT. These results confirm two important hypotheses: $(i)$ it is easier to detect satire if the training and the test samples are gathered from the same publication sources, and $(ii)$ our simple unsupervised domain adaptation method is effective for both data representations. If either hypothesis would have been false, we should not have observed any accuracy improvements for the domain adapted methods. We therefore notice that it is important to split the training and test data by publication source, as we did for FreSaDa, in order to report fair results.

\section{Experiments on News Headlines}
\label{sec_Exp_titles}

Although the reported accuracy rates on full news articles are very high (all of them being over $90\%$), FreSaDa allows us to consider more challenging setups, e.g.~performing satire detection only on the titles. We next present results in this challenging setting.

\begin{figure}[!t]
\begin{center}
\includegraphics[width=0.76\linewidth]{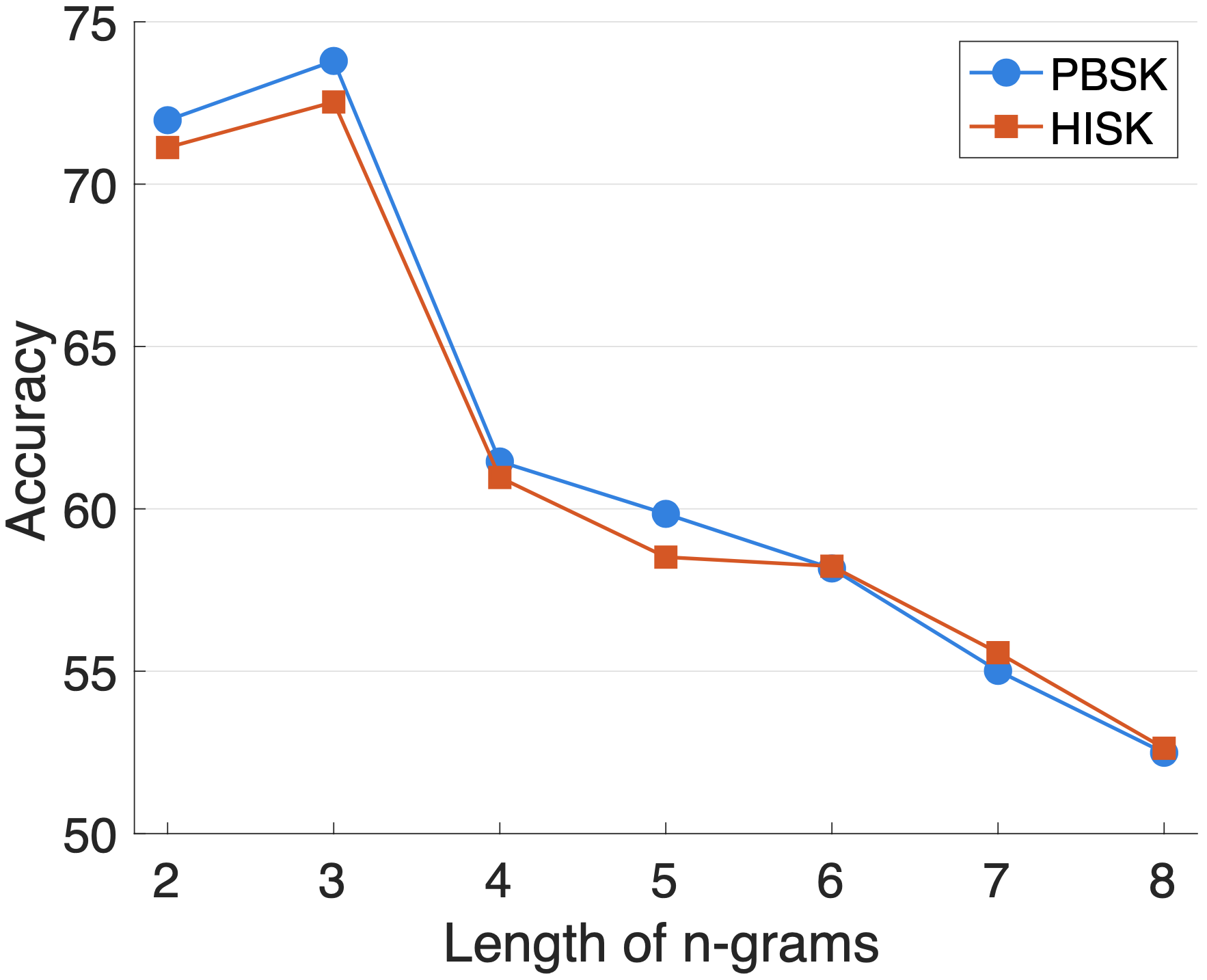}
\end{center}
\caption{Validation accuracy rates on news headlines for PBSK and HISK with character n-grams in the range $\{2,3,4,5,6,7,8\}$, respectively. Best viewed in color.}
\label{fig_ngrams_plot_titles}
\end{figure}

\begin{figure}[!t]
\begin{center}
\includegraphics[width=0.78\linewidth]{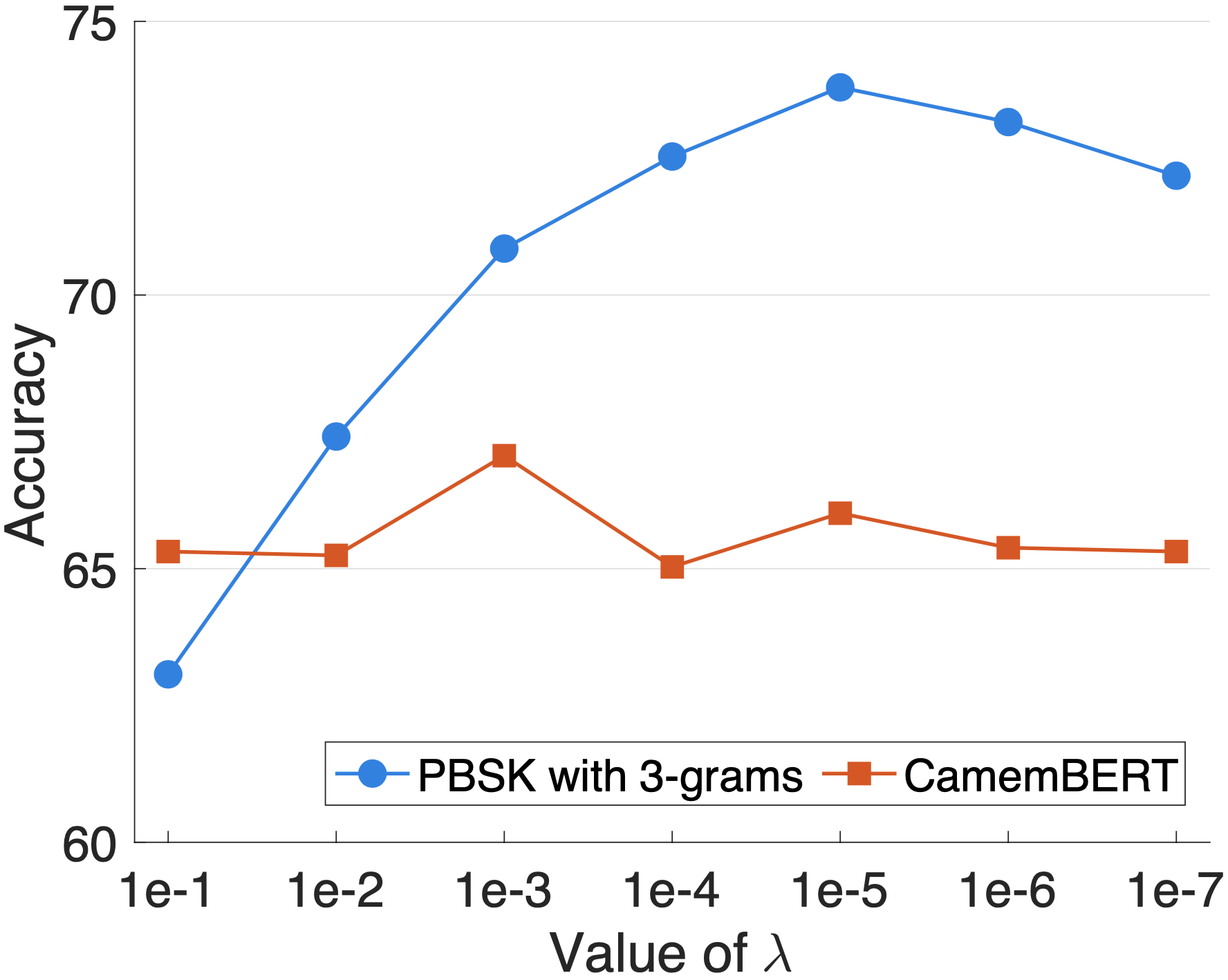}
\end{center}
\caption{Validation accuracy rates on news headlines for PBSK with character 3-grams and for CamemBERT embeddings, with values of $\lambda$ between $10^{-1}$ and $10^{-7}$, where $\lambda$ controls regularization in (Kernel) Ridge Regression. Best viewed in color.}
\label{fig_lambdas_plot_titles}
\end{figure}

\subsection{Hyperparameter Tuning}

As on the full news articles, we observed that (K)RR outperforms SVM. Therefore, all the following experiments are conducted using (K)RR. In order to find the optimal n-gram length for PBSK and HISK, we started with the same range of n-grams ($4$-$8$) as in Section~\ref{sec_Exp_full}. Observing that we obtain superior results with the shortest n-gram length in the range $4$-$8$, we extended the range down to trigrams and bigrams. Hence, the final range of character n-grams for news headlines is $\{2,3,4,5,6,7,8\}$. In Figure~\ref{fig_ngrams_plot_titles}, we present the validation accuracy rates for each n-gram length, using both string kernel methods. We observe that the optimal n-gram length for news headlines is $3$. Since the examples are now significantly shorter in length (compared to the full news articles), the probability of finding a certain n-gram in both training and testing is much lower. Furthermore, the probability drops as the n-gram length gets higher. This explains why results with longer n-grams are worse when the examples are shorter. In the subsequent experiments, we opted for PBSK based on 3-grams. As in Section~\ref{sec_Exp_full}, we next proceed by tuning the regularization parameter $\lambda$ of (K)RR, considering values in the range $\{10^{-1}, 10^{-2}, 10^{-3}, 10^{-4}, 10^{-5}, 10^{-6}, 10^{-7} \}$. In Figure~\ref{fig_lambdas_plot_titles}, we provide the validation accuracy rates for each value of $\lambda$ using either PBSK with 3-grams or the average of CamemBERT word embeddings. For PBSK, we opted for $\lambda=10^{-5}$, while for CamemBERT, we opted for $\lambda=10^{-3}$. When we had to integrate our unsupervised domain adaptation method into a classification model, we repeated the tuning of $\lambda$ for optimal results.

\begin{table}[!t]
\setlength\tabcolsep{5.5pt}
\caption{Validation and test accuracy rates of Ridge Regression using PBSK or CamemBERT representations. Results are reported with and without unsupervised domain adaptation (DA). Results of domain adapted methods marked with $\dagger$ are significantly better than the corresponding baseline, according to a paired McNemar's test \cite{Dietterich-NC-1998} performed at the significance level $0.05$.}
\begin{center}
\begin{tabular}{|l|c|c|}
\hline
Representation 				& Validation            & Test\\
\hline
\hline
PBSK				        & 73.79\%               & 73.86\% \\
CamemBERT			        & 67.06\%		        & 67.13\% \\
\hline
PBSK + DA                   & 74.14\%	            & 74.07\% \\
CamemBERT + DA	            & 73.09\%$^\dagger$		& 72.74\%$^\dagger$ \\  
\hline
\end{tabular}
\end{center}
\label{tab_results_FreSaDa_titles}
\end{table}

\subsection{Results}

Table \ref{tab_results_FreSaDa_titles} contains the validation and the test results of the two baselines (KRR based on PBSK and RR based on CamemBERT embeddings) on news headlines from FreSaDa. Considering the results without domain adaptation, we observe that PSBK significantly outperforms the average of CamemBERT word embeddings. This is likely due to the reduced feature overlap between training and testing, considering the shorter examples and the disjoint publication sources among the two sets of examples. Since PBSK is based on trigrams, which are more likely to appear in both training and testing than words, the model based on character n-grams is able to better cope with the domain gap and the scarce data.

Considering the results with domain adaptation, we notice significant performance improvements for the RR based on CamemBERT embeddings. Domain adaptation seems to bring only minor improvements to PBSK, likely because the accuracy of the model is already saturated. Even though domain adaption has a larger impact for the average of CamemBERT embeddings, PBSK maintains its superiority on news headlines. 

Provided that both methods yield accuracy rates under $75\%$, we conclude that satire detection from news headlines is not an easy task, remaining a difficult challenge to be addressed in future research.

\section{Discriminative Feature Analysis}
\label{sec_Analysis}

We next look at the discriminative features learned by PBSK and CamemBERT. We choose the models without domain adaption in order to analyze the rather more generic features.

\begin{table}[!t]
\caption{Some discriminative features for PBSK, sorted by their score from top to bottom.}
\begin{center}
\begin{tabular}{|l|l|l|l|}
\hline
\multicolumn{2}{|c|}{Satirical}                              & \multicolumn{2}{c|}{Regular}                                    \\ \hline
\multicolumn{1}{|c|}{Feature} & \multicolumn{1}{c|}{Translation} & \multicolumn{1}{c|}{Feature} & \multicolumn{1}{c|}{Translation}     \\ \hline
\hline
reportage                     & report                       & jusqu'                       & until                            \\ \hline
aujourd'hui                   & today                        & national                     & national                         \\ \hline
d'autres                      & other                        & politique                    & political                        \\ \hline
ce matin                      & this morning                 & mercredi                     & Wednesday                        \\ \hline
raconter                      & tell (a story)               & jeudi                        & Thursday                         \\ \hline
jeune                         & young                        & mars                         & March                            \\ \hline
illustration                  & picture                      & en 2012, 2013                & in 2012, 2013              \\ \hline
prochain                      & next                         & international                & international                    \\ \hline
semaine                       & week                         & femmes                       & women                            \\ \hline
l'homme                       & the man                      & millions                     & millions                         \\ \hline
lors d'un/une                 & during an                    & mardi                        & Tuesday                          \\ \hline
affirmer                      & to state                     & notamment                    & notably \\ \hline
\end{tabular}
\end{center}
\label{tab_PBSK_features}
\end{table}

\begin{table}[!t]
\caption{Some discriminative words for CamemBERT, sorted by their score from top to bottom.}
\begin{center}
\begin{tabular}{|l|l|l|l|}
\hline
\multicolumn{2}{|c|}{Satirical}                              & \multicolumn{2}{c|}{Regular}                                \\ \hline
\multicolumn{1}{|c|}{Feature} & \multicolumn{1}{c|}{Translation} & \multicolumn{1}{c|}{Feature} & \multicolumn{1}{c|}{Translation} \\ \hline
\hline
dossier                       & file/folder                  & communiquant                 & communicating                \\ \hline
technologie                   & technology                   & biblioth\'{e}caire               & librarian                    \\ \hline
hamac                         & hammock                      & mexicain                     & Mexican                      \\ \hline
questions                     & questions                    & arm\'{e}nien                     & Armenian                     \\ \hline
imposante                     & impressive                   & producteurs                  & producers                    \\ \hline
mamie                         & grandma                      & parfum                       & perfume                      \\ \hline
baiser                        & kiss/intercourse             & milliardaire                 & billionaire                  \\ \hline
fabulous                      & (imported as is)             & infiniment                   & infinitely                   \\ \hline
d\'{e}bile                        & dummy                        & doyen                        & dean/elder                   \\ \hline
\end{tabular}
\end{center}
\label{tab_CamemBERT_words}
\end{table}

\begin{table*}[!t]
\caption{Some discriminative bigrams for CamemBERT, sorted by their score from top to bottom.}
\begin{center}
\begin{tabular}{|l|l|l|l|}
\hline
\multicolumn{2}{|c|}{Satirical}                              & \multicolumn{2}{c|}{Regular}                                \\ \hline
\multicolumn{1}{|c|}{Word Bigram} & \multicolumn{1}{c|}{Translation} & \multicolumn{1}{c|}{Word Bigram} & \multicolumn{1}{c|}{Translation} \\ \hline
\hline
en \'{e}paules                  & with shoulders                    & moins dans                & less in           \\ \hline
t\^{a}ché malgr\'{e}            & task despite                      & surtout nord              & mainly north      \\ \hline
en cours                        & ongoing                           & voyages provence          & Provence travel   \\ \hline
mani\`{e}re criminelle          & criminal manner                   & tout communiquant         & any communicating \\ \hline
faute m\^{e}me                  & mistake itself                    & Bienvenue Place           & Welcome to Place  \\ \hline
particuli\`{e}rement efficace   & particularly effective            & du mobile                 & of the mobile     \\ \hline
en chansons                     & in songs                          & une plaie                 & a wound           \\ \hline
\'{e}tude mais                  & study but                         & premier protagoniste      & first character   \\ \hline
comp\'{e}tence en               & expertise in                      & printemps 2018            & Spring of 2018    \\ \hline
or noir                         & black gold (metaphor for oil)     & mexicain Carlos           & Carlos the Mexican\\ \hline
en rendit                       & gave something                    & nord am\'{e}ricains       & North Americans   \\ \hline
particuli\`{e}rement souple     & particularly flexible             & un parfum                 & a perfume         \\ \hline
\end{tabular}
\end{center}
\label{tab_CamemBERT_bigrams}
\end{table*}

For PBSK, the most discriminative features for satirical news are related to ``r\'{e}daction'' (``editorial office'') – features ranked on 5-10, ``expliquer'' (``to explain'') – features ranked on 14-16, and ``plusieurs'' (``several'') – features ranked on 26-30. For regular news, the relevant features are ``juillet'' (``July'') – features ranked on 8-9, and ``aujourd'hui'' (``today'') – features ranked on 27-30.
Some other features with their corresponding meaning are listed in Table~\ref{tab_PBSK_features}.
We notice that the features from satirical news are quite opposite in meaning, in comparison with the regular news. For instance, satirical news tend to be vaguer (features meaning ``several'', and so on), while regular news tend to be more precise (features meaning ``million'', and so on). For regular news, PBSK seems to focus on precise temporal aspects, e.g.~``Tuesday'', ``July'', ``2018''. We also observe that many regular news seem to cover international events, while satirical news are more focused on national aspects. Interestingly, we notice one possible bias towards the male/female distinction, since features that mean ``woman'' are weighted higher in the satirical features list, while features that mean ``man'' appear more often in the regular news. 

Since, to the best of our knowledge, there is no agreed standard on identifying the discriminative features for the average of word embeddings, we adopted a simple and straightforward solution. Starting from the assumption that words with higher correlation to the learned weights are more likely to represent the decisions of Ridge Regression based on CamemBERT embeddings than less correlated words, we compute the cosine similarity between each word embedding and the weights learned by Ridge Regression, summing up the results across the entire data set. We employ the same approach for word bigrams, considering that the cosine similarities of every two consecutive words can be multiplied to obtain a representative measure of importance for the corresponding bigrams. We limit ourselves to the analysis of discriminative words and word bigrams, noting that our discriminative feature identification process can be extended to word trigrams and so on.
We acknowledge that our discriminative feature extraction approach is far from being perfect, considering that the most important features of CamemBERT seem to be noisier than those of PBSK. This is most likely because the news articles are represented as an average of CamemBERT word embeddings, generating difficulties in determining the discriminative features.

In Table~\ref{tab_CamemBERT_words}, we present a selection of words relevant for RR based on CamemBERT embeddings. We notice that satirical news are more familiar than regular news in terms of language (``mamie'', meaning ``grandma'', for instance). In satirical news, we observe words with funny connotations, such as ``baiser'', which means ``kiss'' as a noun, but also ``sexual intercourse'', as a verb, in vulgar language. Unlike the regular news, satirical news tend to contain words directly imported from English, such as ``fabulous''. In regular news, one may notice features related to nationalities, e.g.~``Armenian'', ``Mexican'', ``Swedish''. 

In Table~\ref{tab_CamemBERT_bigrams}, we provide a selection of relevant word bigrams. We notice that the bigrams for satirical news still have a vaguer meaning (as for the PBSK features). For instance, the term ``particularly'' appears recurrently. Another recurring term is ``en'', which has several meanings (``in'', ``with'', or a reference to something mentioned before). Another particularity for satirical news is the importance of the feature ``en rendit'' (``gave or surrender something''), for which the verbal time is unusual for this type of writing. This form of past is mostly encountered in novel writing or storytelling. Metaphors are also common in novel writing or poetry. We encountered here a metaphor for ``oil'', namely ``or noir'', meaning ``black gold''. Regarding the regular news, we noticed that some nationality references are still captured by the word bigrams (see Table~\ref{tab_CamemBERT_words}), e.g.~``Carlos the Mexican'' or ``North Americans''. Another recurrent pattern is the use of the indefinite article ``un'' (``a''). References to concrete places (``Provence travel'', ``Bienvenue Place'') or times (``Spring of 2018'') can also be noticed in the French word bigrams relevant for regular news. 

\section{Conclusion}
\label{sec_Conclusion}

In this paper, we presented a novel and large corpus of French satirical and regular news. On this corpus, we proposed two cross-domain (cross-source) satire detection tasks, one considering full news articles and another considering only news headlines. As baselines, we employed two classification methods, one based on low-level features (PBSK) and one based on high-level features (CamemBERT). Another contribution introduced in our work is an unsupervised domain adaptation method based on regarding the pairwise similarities between the training samples and the validation samples as features. By including domain-specific features, we attained statistically significant improvements for both PBSK and CamemBERT. In our work, we also discussed the most important features of both machine learning models. 

After experimenting on the test set of full news articles, we achieved the top accuracy rate of $97.48\%$. We observed that satire detection on news headlines is significantly more challenging, the top accuracy rate being $74.07\%$. We therefore conclude that the latter task is quite challenging, being an open problem to be addressed in future research. 

In future work, we plan to enlarge the corpus and to diversify the tasks, perhaps by considering the collection of the associated images for each news article, which would give rise to a multi-modal satire detection task.

\section*{Acknowledgment}
This work was supported by a grant of the Romanian Ministry of Education and Research, CNCS - UEFISCDI, project number PN-III-P1-1.1-TE-2019-0235, within PNCDI III. This article has also benefited from the support of the Romanian Young Academy, which is funded by Stiftung Mercator and the Alexander von Humboldt Foundation for the period 2020-2022.



%
\bibliographystyle{IEEEtran}
\bibliography{references}

\begin{thebibliography}{10}
\providecommand{\url}[1]{#1}
\csname url@samestyle\endcsname
\providecommand{\newblock}{\relax}
\providecommand{\bibinfo}[2]{#2}
\providecommand{\BIBentrySTDinterwordspacing}{\spaceskip=0pt\relax}
\providecommand{\BIBentryALTinterwordstretchfactor}{4}
\providecommand{\BIBentryALTinterwordspacing}{\spaceskip=\fontdimen2\font plus
\BIBentryALTinterwordstretchfactor\fontdimen3\font minus
  \fontdimen4\font\relax}
\providecommand{\BIBforeignlanguage}[2]{{%
\expandafter\ifx\csname l@#1\endcsname\relax
\typeout{** WARNING: IEEEtran.bst: No hyphenation pattern has been}%
\typeout{** loaded for the language `#1'. Using the pattern for}%
\typeout{** the default language instead.}%
\else
\language=\csname l@#1\endcsname
\fi
#2}}
\providecommand{\BIBdecl}{\relax}
\BIBdecl

\bibitem{Reganti-ICDMW-2016}
A.~N. Reganti, T.~Maheshwari, U.~Kumar, A.~Das, and R.~Bajpai, ``{Modeling
  Satire in English Text for Automatic Detection},'' in \emph{Proceedings of
  ICDMW}, 2016, pp. 970--977.

\bibitem{McHardy-NAACL-2019}
R.~McHardy, H.~Adel, and R.~Klinger, ``{Adversarial Training for Satire
  Detection: Controlling for Confounding Variables},'' in \emph{Proceedings of
  NAACL}, 2019, pp. 660--665.

\bibitem{Barbieri-PLN-2015}
F.~Barbieri, F.~Ronzano, and H.~Saggion, ``{Is this tweet satirical? A
  computational approach for satire detection in Spanish},''
  \emph{Procesamiento de Lenguaje Natural}, vol.~55, pp. 135--142, 2015.

\bibitem{Karoui2017}
J.~Karoui, F.~Zitoune, and V.~Moriceau, ``{SOUKHRIA: Towards an Irony Detection
  System for Arabic in Social Media},'' in \emph{Proceedings of ACLing}, vol.
  117, 12 2017, pp. 161--168.

\bibitem{Liebrecht2013}
C.~Liebrecht, F.~Kunneman, and A.~van~den Bosch, ``The perfect solution for
  detecting sarcasm in tweets {\#}not,'' in \emph{Proceedings of WASSA}, 2013,
  pp. 29--37.

\bibitem{Jia2019}
X.~Jia, Z.~Deng, F.~Min, and D.~Liu, ``{Three-way decisions based feature
  fusion for Chinese irony detection},'' \emph{International Journal of
  Approximate Reasoning}, vol. 113, pp. 324--335, 2019.

\bibitem{Giudice-EVALITA-2018}
V.~Giudice, ``{Aspie96 at IronITA (EVALITA 2018): Irony Detection in Italian
  Tweets with Character-Level Convolutional RNN},'' in \emph{Proceedings of
  EVALITA}, 2018, pp. 160--165.

\bibitem{Meel-ESA-2019}
P.~Meel and D.~K. Vishwakarma, ``{Fake news, rumor, information pollution in
  social media and web: A contemporary survey of state-of-the-arts, challenges
  and opportunities},'' \emph{Expert Systems with Applications}, vol. 153, p.
  112986, 2019.

\bibitem{Perez-COLING-2018}
V.~P{\'e}rez-Rosas, B.~Kleinberg, A.~Lefevre, and R.~Mihalcea, ``{Automatic
  Detection of Fake News},'' in \emph{Proceedings of COLING}, 2018, pp.
  3391--3401.

\bibitem{Sharma-TIST-2019}
K.~Sharma, F.~Qian, H.~Jiang, N.~Ruchansky, M.~Zhang, and Y.~Liu, ``{Combating
  fake news: A survey on identification and mitigation techniques},'' \emph{ACM
  Transactions on Intelligent Systems and Technology}, vol.~10, no.~3, pp.
  1--42, 2019.

\bibitem{Martin-Arxiv-2019}
L.~Martin, B.~Muller, P.~J. Ortiz~Su{\'a}rez, Y.~Dupont, L.~Romary,
  {\'E}.~de~la Clergerie, D.~Seddah, and B.~Sagot, ``{CamemBERT: a Tasty French
  Language Model},'' in \emph{Proceedings of ACL}, 2020, pp. 7203--7219.

\bibitem{Burfoot-ACL-2009}
C.~Burfoot and T.~Baldwin, ``{Automatic Satire Detection: Are You Having a
  Laugh?}'' in \emph{Proceedings of ACL-IJCNLP}, 2009, pp. 161--164.

\bibitem{Frain-LREC-2016}
A.~Frain and S.~Wubben, ``{SatiricLR: a Language Resource of Satirical News
  Articles},'' in \emph{Proceedings of LREC}, 2016, pp. 4137--4140.

\bibitem{Goldwasser-TACL-2016}
D.~Goldwasser and X.~Zhang, ``{Understanding Satirical Articles Using
  Common-Sense},'' \emph{Transactions of the Association for Computational
  Linguistics}, vol.~4, pp. 537--549, 2016.

\bibitem{Li-NLP4IF-2020}
L.~Li, O.~Levi, P.~Hosseini, and D.~Broniatowski, ``{A Multi-Modal Method for
  Satire Detection using Textual and Visual Cues},'' in \emph{Proceedings of
  NLP4IF}, 2020, pp. 33--38.

\bibitem{Lu2019}
J.~Lu, D.~Batra, D.~Parikh, and S.~Lee, ``{ViLBERT: Pretraining Task-Agnostic
  Visiolinguistic Representations for Vision-and-Language Tasks},'' in
  \emph{Proceedings of NeurIPS}, vol.~32, 2019, pp. 13--23.

\bibitem{Ravi-KBS-2017}
K.~Ravi and V.~Ravi, ``A novel automatic satire and irony detection using
  ensembled feature selection and data mining,'' \emph{Knowledge-Based
  Systems}, vol. 120, pp. 15--33, 2017.

\bibitem{Yang-EMNLP-2017}
F.~Yang, A.~Mukherjee, and E.~Dragut, ``{Satirical News Detection and Analysis
  using Attention Mechanism and Linguistic Features},'' in \emph{Proceedings
  EMNLP}, 2017, pp. 1979--1989.

\bibitem{Saadany-RDSM-2020}
H.~Saadany, C.~Orasan, and E.~Mohamed, ``{Fake or Real? A Study of Arabic
  Satirical Fake News},'' in \emph{Proceedings of RDSM}, 2020, pp. 70--80.

\bibitem{Tocoglu-ID-2019}
M.~A. To{\c{c}}o{\u{g}}lu and A.~Onan, ``{Satire Detection in Turkish News
  Articles: A Machine Learning Approach},'' in \emph{Proceedings of
  Innovate-Data}, 2019, pp. 107--117.

\bibitem{Guibon-CICLing-2019}
G.~Guibon, L.~Ermakova, H.~Seffih, A.~Firsov, G.~Le~Noé-Bienvenu, and
  G.~Guibon, ``{Multilingual Fake News Detection with Satire},'' in
  \emph{Proceedings of CICLing}, 2019.

\bibitem{Liu-ICCCN-2019}
Z.~Liu, S.~Shabani, N.~G. Balet, and M.~Sokhn, ``{Detection of Satiric News on
  Social Media: Analysis of the Phenomenon with a French Dataset},'' in
  \emph{Proceedings of ICCCN}, 2019, pp. 1--6.

\bibitem{Butnaru-VarDial-2018}
A.~M. Butnaru and R.~T. Ionescu, ``{UnibucKernel Reloaded: First Place in
  Arabic Dialect Identification for the Second Year in a Row},'' in
  \emph{Proceedings of VarDial}, 2018, pp. 77--87.

\bibitem{Cozma-ACL-2018}
M.~Cozma, A.~Butnaru, and R.~T. Ionescu, ``Automated essay scoring with string
  kernels and word embeddings,'' in \emph{Proceedings of ACL}, 2018, pp.
  503--509.

\bibitem{Ionescu-VarDial-2017}
R.~T. Ionescu and A.~M. Butnaru, ``{Learning to Identify Arabic and German
  Dialects using Multiple Kernels},'' in \emph{Proceedings of VarDial}, 2017,
  pp. 200--209.

\bibitem{Ionescu-EMNLP-2014}
R.~T. Ionescu, M.~Popescu, and A.~Cahill, ``{Can characters reveal your native
  language? A language-independent approach to native language
  identification},'' in \emph{Proceedings of EMNLP}, 2014, pp. 1363--1373.

\bibitem{Gimenez-EACL-2017}
R.~M. Gim\'{e}nez-P\'{e}rez, M.~Franco-Salvador, and P.~Rosso, ``{Single and
  Cross-domain Polarity Classification using String Kernels},'' in
  \emph{Proceedings of EACL}, 2017, pp. 558--563.

\bibitem{Ionescu-COLI-2016}
R.~T. Ionescu, M.~Popescu, and A.~Cahill, ``String kernels for native language
  identification: Insights from behind the curtains,'' \emph{Computational
  Linguistics}, vol.~42, no.~3, pp. 491--525, 2016.

\bibitem{Popescu-BEA8-2013}
M.~Popescu and R.~T. Ionescu, ``{The Story of the Characters, the DNA and the
  Native Language},'' in \emph{Proceedings of BEA-8}, 2013, pp. 270--278.

\bibitem{Devlin-NAACL-2019}
J.~Devlin, M.-W. Chang, K.~Lee, and K.~Toutanova, ``{BERT: Pre-training of Deep
  Bidirectional Transformers for Language Understanding},'' in
  \emph{Proceedings of NAACL}, 2019, pp. 4171--4186.

\bibitem{Liu-Arxiv-2019}
Y.~Liu, M.~Ott, N.~Goyal, J.~Du, M.~Joshi, D.~Chen, O.~Levy, M.~Lewis,
  L.~Zettlemoyer, and V.~Stoyanov, ``{RoBERTa: A Robustly Optimized BERT
  Pretraining Approach},'' \emph{arXiv preprint arXiv:1907.11692}, 2019.

\bibitem{Suarez-CMLC-2019}
P.~J.~O. Su{\'a}rez, B.~Sagot, and L.~Romary, ``{Asynchronous Pipeline for
  Processing Huge Corpora on Medium to Low Resource Infrastructures},'' in
  \emph{Proceedings of CMLC-7}, 2019, pp. 9--16.

\bibitem{Vaswani-NIPS-2017}
A.~Vaswani, N.~Shazeer, N.~Parmar, J.~Uszkoreit, L.~Jones, A.~N. Gomez,
  {\L}.~Kaiser, and I.~Polosukhin, ``Attention is all you need,'' in
  \emph{Proceedings of NIPS}, 2017, pp. 5998--6008.

\bibitem{Fu-ESA-2018}
M.~Fu, H.~Qu, L.~Huang, and L.~Lu, ``{Bag of meta-words: A novel method to
  represent document for the sentiment classification},'' \emph{Expert Systems
  with Applications}, vol. 113, pp. 33--43, 2018.

\bibitem{Ionescu-KES-2017}
A.~Butnaru and R.~T. Ionescu, ``{From Image to Text Classification: A Novel
  Approach based on Clustering Word Embeddings},'' in \emph{Proceedings of
  KES}, 2017, pp. 1784--1793.

\bibitem{Ionescu-NAACL-2019}
R.~T. Ionescu and A.~Butnaru, ``{Vector of Locally-Aggregated Word Embeddings
  (VLAWE): A Novel Document-level Representation},'' in \emph{Proceedings of
  NAACL}, 2019, pp. 363--369.

\bibitem{Shen-ACL-2018}
D.~Shen, G.~Wang, W.~Wang, M.~R. Min, Q.~Su, Y.~Zhang, C.~Li, R.~Henao, and
  L.~Carin, ``{Baseline Needs More Love: On Simple Word-Embedding-Based Models
  and Associated Pooling Mechanisms},'' in \emph{Proceedings of ACL}, 2018, pp.
  440--450.

\bibitem{Cortes-ML-1995}
C.~Cortes and V.~Vapnik, ``{Support-Vector Networks},'' \emph{Machine
  Learning}, vol.~20, no.~3, pp. 273--297, 1995.

\bibitem{Hoerl-Technometrics-1970}
A.~E. Hoerl and R.~W. Kennard, ``{Ridge Regression: Biased estimation for
  nonorthogonal problems},'' \emph{Technometrics}, vol.~12, no.~1, pp. 55--67,
  1970.

\bibitem{Saunders-ICML-1998}
C.~Saunders, A.~Gammerman, and V.~Vovk, ``{Ridge Regression Learning Algorithm
  in Dual Variables},'' in \emph{Proceedings of ICML}, 1998, pp. 512--521.

\bibitem{Dietterich-NC-1998}
T.~G. Dietterich, ``{Approximate Statistical Tests for Comparing Supervised
  Classification Learning Algorithms},'' \emph{Neural Computation}, vol.~10,
  no.~7, pp. 1895--1923, 1998.

\end{thebibliography}

\end{document}